\documentclass[10pt,letterpaper]{article}
\usepackage[top=0.85in,left=2.75in,footskip=0.75in]{geometry}

\usepackage{amsmath,amssymb}

\usepackage{changepage}

\usepackage[utf8x]{inputenc}

\usepackage{textcomp,marvosym}

\usepackage{cite}

\usepackage{nameref,hyperref}

\usepackage{microtype}
\DisableLigatures[f]{encoding = *, family = * }

\usepackage[table]{xcolor}

\usepackage{array}
\usepackage{graphicx}

\usepackage[numbers]{natbib}
\usepackage{multirow}
\usepackage{subcaption}
\usepackage{amsmath}
\usepackage{float}
\usepackage{amsthm}
\usepackage{placeins}
\usepackage{algpseudocode}

\usepackage{algorithm}
\renewcommand{\algorithmicrequire}{\textbf{Input:}}
\renewcommand{\algorithmicensure}{\textbf{Output:}}
\def\tsc#1{\csdef{#1}{\textsc{\lowercase{#1}}\xspace}}
\tsc{WGM}
\tsc{QE}
\tsc{EP}
\tsc{PMS}
\tsc{BEC}
\tsc{DE}

\newcolumntype{+}{!{\vrule width 2pt}}

\newlength\savedwidth

\raggedright
\usepackage[aboveskip=1pt,labelfont=bf,labelsep=period,justification=raggedright,singlelinecheck=off]{caption}

\makeatletter
\renewcommand{\@biblabel}[1]{\quad#1.}
\makeatother

\usepackage{lastpage,fancyhdr,graphicx}
\usepackage{epstopdf}
\fancyheadoffset[L]{2.25in}
\fancyfootoffset[L]{2.25in}
\begin{document}
\vspace*{0.2in}

\begin{flushleft}
{\Large
\textbf\newline{InstructNet: A Novel Approach for Multi-Label Instruction Classification through Advanced Deep Learning} 
}
\newline
\\
Tanjim Taharat Aurpa\textsuperscript{1,2},
Md Shoaib Ahmed\textsuperscript{1,3},
Md Mahbubur Rahman\textsuperscript{1,4},
Md. Golam Moazzam\textsuperscript{1}\\
\bigskip
\textbf{1} Department of Computer Science and Engineering, Jahangirnagar University, Savar, Dhaka
\\
\textbf{2} Department of Data Science and Engineering, Bangabandhu Sheikh Mujibur Rahman Digital University, Gazipur, Bangladesh
\\
\textbf{3} Department of Computer Science, Boise State University, Boise, ID, USA
\\
\textbf{4} Iowa State University, Iowa, USA
\\
\bigskip

%
%

* Corresponding Author 

\end{flushleft}
\begin{abstract}
People use search engines for various topics and items, from daily essentials to more aspirational and specialized objects. Therefore, search engines have taken over as people's preferred resource. The "How To" prefix has become familiar and widely used in various search styles to find solutions to particular problems. This search allows people to find sequential instructions by providing detailed guidelines to accomplish specific tasks. Categorizing instructional text is also essential for task-oriented learning and creating knowledge bases.
This study uses the "How To" articles to determine the multi-label instruction category. We have brought this work with a dataset comprising 11,121 observations from wikiHow, where each record has multiple categories. To find out the multi-label category meticulously, we employ some transformer-based deep neural architectures, such as Generalized Autoregressive Pretraining for Language Understanding (XLNet), Bidirectional Encoder Representation from Transformers (BERT), etc.
In our multi-label instruction classification process, we have reckoned our proposed architectures using accuracy and macro f1-score as the performance metrics. This thorough evaluation showed us much about our strategy's strengths and drawbacks. Specifically, our implementation of the XLNet architecture has demonstrated unprecedented performance, achieving an accuracy of 97.30\% and micro and macro average scores of 89.02\% and 93\%, a noteworthy accomplishment in multi-label classification. This high level of accuracy and macro average score is a testament to the effectiveness of the XLNet architecture in our proposed 'InstructNet' approach. By employing a multi-level strategy in our evaluation process, we have gained a more comprehensive knowledge of the effectiveness of our proposed architectures and identified areas for forthcoming improvement and refinement.           
\end{abstract}

\section{Introduction}

Instructions and guidelines are mandatory for any well-defined task, as they help anyone execute work independently. Instructions or guidelines are generalized lists of steps that enable anyone to understand the working principle. Instead of listening to work procedures, modern people prefer to search for instructions on Google. It is necessary to categorize instructions efficiently to ensure an efficient search experience. Furthermore, one instruction can be employed in multiple categories.
Implementing a solution that can help humankind analyze systems that are able to categorize instructions automatically can be revolutionary for searching. 
Additionally, the wikiHow articles are used for task-oriented learning to create knowledge bases, intelligent agents, chatbots, graph creation, etc., where step-by-step guidelines are learned by these systems. During these types of automated intelligent implementations, classifying wikiHow articles and tagging them with multiple appropriate tags can be very beneficial. It can help to get the answers to questions such as 
"In which the instructional text learned by the intelligent system belongs?" or 
"Can one learned instruction can be used for multiple contexts?"
Nevertheless, no significant work has been conducted on the research topic of multilabel instructional text classification, which requires significant solutions.

The research attention in this area is not considered, whereas solutions such as wikiHow (https://www.wikihow.com/Main-Page) generate various instructions daily. 
WikiHow contains instructions in the "How to" manner for various tasks. The database is enriched with visual contents of instructions, which have all gone through the expert's verification. This wiki-style website has 2.5 million registered users and has already featured more than  235,000 how-to articles till December 2021. These articles are categorized into different classes. Some of these can even be labeled under multiple categories or tags. Automatically identifying those tags can be an efficient way to search for guidelines. 
These wikiHow instructions are primarily focused on summarization tasks. Moreover, these instructions are shown to contribute to procedural-oriented task learning and to creating a knowledge base. Many datasets are created based on wikiHow articles. Besides learning task-oriented steps and creating a knowledge base, it is important to learn appropriate tags for the instructional texts. Moreover, Classifying them is also necessary because people need more time to accomplish any task. They google the procedure for their required task and follow it. Again, a single instruction can be classified into multiple categories.

Transformer architectures are now prevalent in natural language processing-related research.
These models use the encoder-decoder approach, where the encoder encodes the input to the context vector, and the decoder does the opposite of it. The self-attention mechanism in the transformer provides significant results and contributes to developing various Large Language Models(LLM), such as BERT, ELECTRA, XLNet, etc. All of these models are pre-trained in a self-supervised fashion on a large amount of raw text; as a result, they show remarkable performance in the downstream tasks.
The transformer-based architecture BERT (Bidirectional Encoder Representations from Transformers) brought a revolution in research related to natural language processing. It trains the transformer bidirectionally and uses positional encoding for sequences. This model works in two steps- Pretraining and fine-tuning.
BERT uses two pretraining tasks, Mask Language Modeling(MLM) and Next Sentence Prediction(NSP), known as autoencoding models. These pre-trained models can then be fine-tuned for different downstream Tasks such as classification, Question Answering \cite{aurpa2022reading, aurpa2023uddipok}, etc.
It provided noteworthy performance in various NLP tasks like text classification \cite{aurpa2022progressive, kulkarni2021experimental, krishnan2021cross}, offensive language detection \cite{colla2020grupato}, Entity Relation and Recognition \cite{xue2019fine, aurpa2024ensemble} Translation Quality Estimation \cite {chowdhury2021ensemble, gonen2020s}. 

XLNet is a popular generalized auto-regressive transformer that has shown its performance in various NLP tasks, such as Name Entity Recognition \cite{yan2021named},
Sentiment Analysis \cite{sweidan2021sentence}, Emotion detection \cite{shen2021dialogxl} \cite{adoma2020comparative} etc. 
XLNet is a lengthened version of transformer-XL, which abolishes the limitation of transformers regarding the text sequence length. 
XLNet combines the concept of autoencoding models and autoregressive language models while ignoring their drawbacks. Bidirectional context and positional encoding are both maintained in XLNet architecture.
It uses a different attention mechanism, namely Two stream self-attention.
Like BERT, XLNet also has two steps: pretraining and fine-tuning.
It uses Permutation Language Modeling (PLM) as the pretraining task and considers all the permutations of a sequence. It overcomes the limitation of BERT and shows a significant performance for the downstream tasks. XLNet outperformed BERT in twenty different Natural Language Processing Tasks.

Besides, variants of BERT, RoBERTa, AlBERT, and DistilBERT are utilized in NLP tasks nowadays. Another transformer architecture, ELECTRA, has been attracting researchers recently. Research work such as sentiment/emotion analysis (e.g., \cite{xu2020dombert}, \cite{al2021evolution}), text mining (e.g., \cite{ozyurt2020effectiveness}, fake news analysis (e.g., \cite{das2020ensemble}). All of these transformer models are very similar to the XLNet and BERT. We also utilized them in our data to justify the state-of-the-art performance of our proposed method.

\subsection{Research Objective}

In this research, we have proposed an autoregressive transformer-based architecture for multilabel instruction classification. To the best of our knowledge, this is the first work for multilabel instruction classification using wikiHow articles. Previously, we have exclusively handled instructions that were categorized as single-category multiclass, where the instructions we have dealt with can only be classified into one category \cite{aurpa2022progressive}.

The main contribution of our work has been mentioned below:

\begin{itemize}
\item  Our work Introduces a groundbreaking transformer-based architecture, leveraging XLNet, to effectively categorize wikiHow articles for multilabel instructional text classification. This is the first-ever endeavor in multilabel instruction classification, marking a significant advancement in the field.
\item Our research also addresses the crucial aspect of data preparation for multilabel classification. We propose an algorithm that efficiently filters labels based on a given score, enhancing the practicality and applicability of our methodology.
 \item Our methodology is rigorously evaluated using appropriate metrics, namely Accuracy and Macro F1 Score. This comprehensive evaluation provides robust evidence for the state-of-the-art performance of our proposed methods.
\item Creating a comparison scenario with other existing Large Language Models and visualizing the noteworthy performance.
\end{itemize}
We named the approach 'InstructNet' because we are emphasizing Instructional Text here. In the proposed approach, the primary classifier is XLNet. Therefore, the term 'InstructNet' is created by merging Instruction and XLNet. 

Section 2, named Relates Works, contains the required literature review for this work. We have provided the preliminary concept and the overview of the proposed method in Section 3. In Section 4, we have given the pre-experimental setup. Then, Section 5 contains all of our findings and results. Lastly, Section 6 discusses our research.

\section{Related Work}

wikiHow instructions have been utilized as research data in several deep-learning works. However, most of those are conducted for summarization tasks. 
Jadeja et al. \cite{jadeja2022comparative} used two models, PEGASUS and T5, for text summarization tasks with the wikiHow dataset. They compared the proposed methods based on evolution metrics (BLEU and ROUGE) and human opinion. 
For lengthy text summarization, a sentence filtering technique is proposed in the paper \cite{mei2022learning}. It removes the low-quality sentences and preserves the highly informative ones.
Authors in \cite{srivastava2022topic} proposed an unsupervised topic modeling approach for text summarization, combining Latent Dirichlet Allocation (LDA) and K-Medoids clustering. 
Other types of tasks on wikiHow instructions are learning different kinds of knowledge. Zhou showed a procedural knowledge-acquiring method from household instructions \cite{zhou2019learning}, and Zhang proposed a technique for reasoning tasks for relating goals and steps.
In the paper \cite{devi2023text} wikiHow, articles are used for both text summarization and categorization. 

wikiHow instructions have been less utilized for text classification tasks; even the existing classification problems are only partially focused on text. Lin \cite{lin2022learning} et al. proposed a method to classify multistep activities from lengthy video spanning. They used a distant supervision technique that recognized the procedure steps from the wikiHow instructional video. 
Multi-task classification on wikiHow articles has been executed in \cite{nouriborji2022nowruz} where authors used different transformer models and ordinal regression. DeBERTa\-$\text{V3}_{large}$ has provided the best result with an accuracy score of 64.12\%.
Another work \cite{wiriyathammabhum2021ttcb} on the wikiHow article was done to determine whether an article need revision or not. They achieved the highest 68.84\% accuracy using the XLNet.
Another work \cite{mueller2022label} used a few shot classification models for the Label Semantics Aware Pre-training (LASP), which is a very important task to improve text classification. To supplement their data, they have added the wikiHow intent dataset \cite{zhang2020intent}. 
Aurpa et al. \cite{aurpa2022progressive} use the BERT model to classify wikiHow summary classification. 
None of these works considered the multilabel classification for wikiHow text.

XLNet has been used broadly in text classification tasks. Wang et al. \cite{wang2022performance} proposed an improved version of XLNet for text classification. Authors here showed almost 94.57\% accuracy and 0.3133 loss rate. 

Another contribution of XLNet can be seen in \cite{salma2021text}, where authors use this model for auto-labeling for text classification. They achieved an accuracy of almost 88.68
Authors in \cite{wang2021xlnet} used XLNet for personality classification from textual data. They identified the micro and macro F1 scores for five big models using the XLNet. Liu proposed an XLNet-based hospitality and tourism-related text classification method in \cite{liu2023text}. XLNet outperformed here with 72.2\% accuracy in the Yelp dataset. XLNet scored 96\% and performed best among different classifiers, including the transformer models for text classification in \cite{arabadzhieva2022comparison}.
The use of BERT is also prevalent in text classification tasks. Authors in \cite{li2019automatic} propose a BERT CNN-based model and improve the accuracy of existing works. Another text classification approach is proposed in \cite{yu2021research} where BERT and BiGRU models are combined for the Chinese text classification. This method performed more .9 accuracy and F1 score. Another work \cite{chen2022long} combined BERT with CNN for long Chinese text classification and provided improved accuracy.

XLNet showed better results in multilabel text classifications. 
Roudsari introduced PatentNet in \cite{haghighian2022patentnet} for the multilabel classification of patent documents with various deep-learning models. Authors here also achieved the best performance using the XLNet. Emotion classification with the multilabel approach was done in \cite{ameer2023multi} and reached the highest 45.6\% accuracy with XLNet-MA. In the same way as XLNet, BERT also showed its popularity in multilabel text classification. Chalkidis et al. \cite{chalkidis2019large} used BERT for large-scale multilabel classification. Another work \cite{zhang2021multi} has conducted multilabel text classification using biomedical texts. They used BERT in their proposed method and recognized aspect categories. A combined BERT model for tagging documents is proposed in \cite{cai2020hybrid}. They utilized not only the label semantics but also the fine-grained information of the text. These works are excellent research, undoubtedly.
However, the use of XLNet on wikiHow data for multiple tag recommendation (multilabel classification) has yet to be considered. This is the primary intention of our research. Powerful large language models like BERT XLNET have yet to be used in the classification or tagging of instructional text, such as wikiHow articles.

\section{Preliminary and Proposed Framework}

\subsection{Preliminary Concept}
\subsubsection{Transformers and Transformer-XL model}

The popular Natural Language Processing architecture Transformers \cite{vaswani2017attention}, which is also called encoded decoded models, gained popularity for its multi-head attention mechanism. The encoder in transformer models deals with the sequential input and maps input $(x_1, ..., x_n)$ into the context representation $z (z_1, ..., z_n)$. The decoder utilized this context for the development of output representation $(y_1, ..., y_m)$. The encoder-decoder stacks of the transformer accommodate encoder and decoder layers with $N=6$ layers and two sublayers. These sublayers come about self-attention and fully connected position-wise feed-forward layers. There is a resultant $d_{model} = 512$ dimensional sublayer Layernorm(x+sublayer(x)). 

\begin{figure*}
    \centering
    \includegraphics[width=.8\linewidth]
    {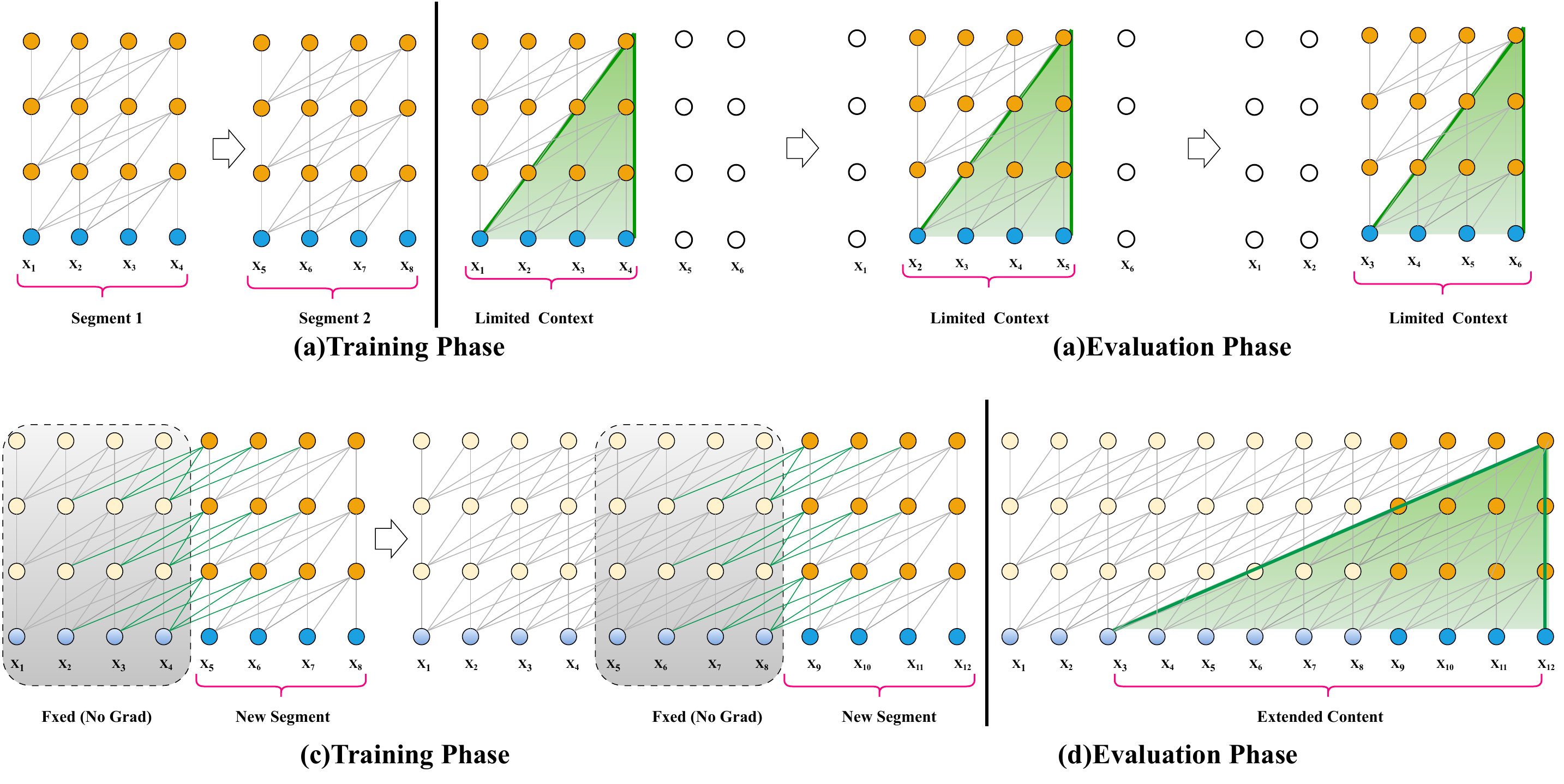}
    \caption{Illustration of How Transformer-XL works with long sequences.(If we compare the figures of the transformer and transformer XL, we can observe that the context representation is limited in transformers where the transformer XL is able to represent long sequences more efficiently.)}
    \label{fig:transXL}
\end{figure*}

Transformer-XL is a neural architecture that combines Transformers and RNNs to improve language modeling. It proposes a segment-level recurrence mechanism that uses concealed states from earlier segments as an extended context for the present element, as shown in Figure \ref{fig:transXL}. This allows learning dependency to extend beyond a defined length without interfering with temporal coherence. As a result, it surpasses RNNs and vanilla Transformers, which are longer than 80\% and 450\%, respectively, on eclectic word-level and character-level datasets and can generate coherent long-text articles. It can measure the ability to model long-term dependency through a new metric called Relative Effective Context Length (RECL) \cite{dai2019transformer}. 

Figure \ref{fig:transXL} (a) and (b) shows the Training and Evaluation phases of normal transformers. Figure \ref{fig:transXL} (c) and (d) is about the Training and Evaluation phases of Transformer XL.

\subsubsection{Bidirectional Encoder Representation from Transformers (BERT)}
\begin{figure}
    \centering
    \includegraphics[width=.7\linewidth]{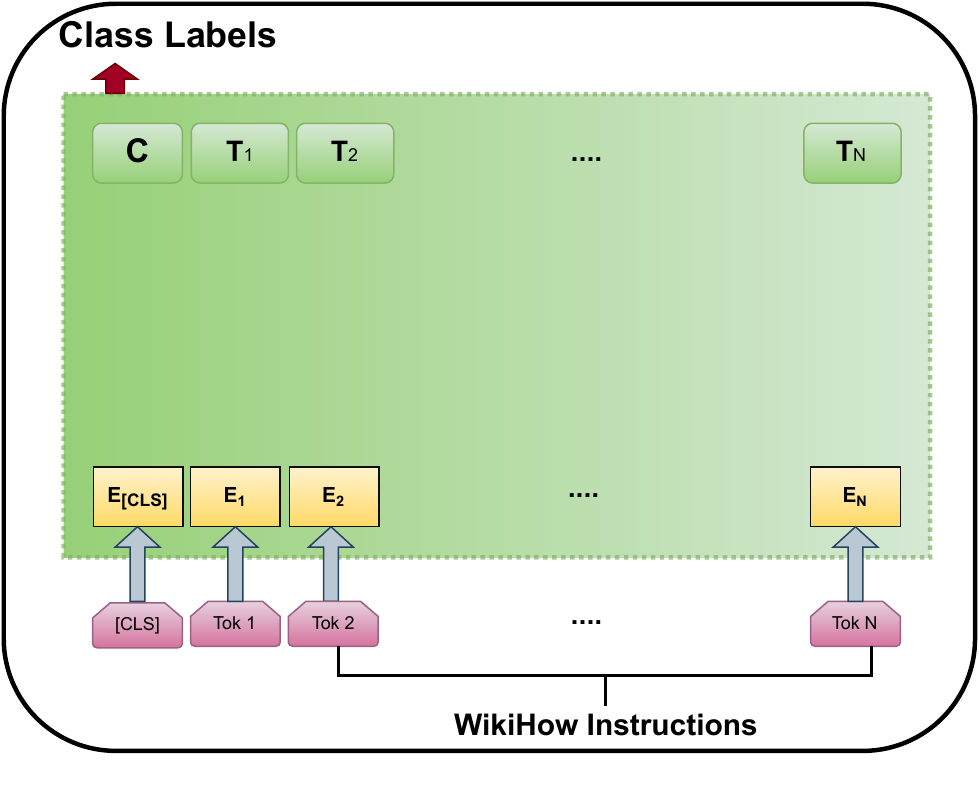}
    \caption{The training phase of the BERT model. The wikiHow instructions are tokenized after combining BERT's special tokens [CLS].}
    \label{fig:bert}
\end{figure}

According to Devlin \cite{devlin2018bert}, BERT is a multilayered bidirectional transformer encoder. An exact token sequence that may incorporate one or more sentences is used as the input for BERT.
BERT consists of the following two steps:

\begin{itemize}
\item\textbf{Pretraining BERT: } The two unsupervised tasks, Masked LM (Language Model) and NSP (Next Sentence Prediction), are utilized to pre\-train BERT. To obtain the pre-trained bidirectional model, masked LM involves masking an array of random tokens and making predictions about them. In this approach, 15\% input tokens are randomly masked, and the model attempts to regenerate original inputs while learning weights.
The goal of NSP is to anticipate the subsequent sentence in a sentence pair. BERT takes this pair and aims to predict whether the sentences in the pair are subsequent sentences or not. 
BERT has been pre-trained using English Wikipedia's text passages (not lists, headings, or Tables) (2,500M words) and BooksCorpus (800M words) \cite{zhu2015aligning}. This is how BERT utilized the self-supervised mechanism. Figure \ref{fig:bert} represents the pretraining task of BERT.

\item \textbf{Fine-tuning BERT: }  BERT is recognized for several downstream functions, including the ability to select appropriate inputs for both single and coupled sentences. It starts with pre-trained parameters, which can be adjusted further in later jobs using labeled data. The pre-trained BERT model shows noteworthy performance in popular downstream tasks, such as question answering, text classification, named entity recognition, machine translation, etc. In this work, we fine-tuned BERT for multilabel text classification.

\end{itemize}

\subsubsection{XLNet}
XLNet\cite{yang2019xlnet} is an autoregressive transformer model that outperforms BERT in various NLP tasks. It utilizes the advantages of the Autoregressive(AR) models while preserving the bidirectional context like an Autoencoded(AE) model. This is a broad version of the Transformer-XL model. It utilizes the pretraining process of Transformer XL and preserves the advantages of AE language models.

For given any text sequence $x = {x_1, x_2, ........, x_T} $ autoregressive models are pre-trained with the following factorization in Equation \ref{eq:AR model}:
\begin{equation}
\begin{split}
\label{eq:AR model}
    \underset{\theta}{\mathrm{max}}\quad log\ p_\theta(x)=\sum_{t=1}^{T} log\ p_\theta(x_t|\mathbf{x}_{<t}) 
= \sum_{t=1}^{T} log\frac{\mathrm{exp}(h_\theta(x_{1:t-1})^\textrm{T}e(x_t))}{\sum_{{x}'}\mathrm{exp}(h_\theta(x_{1:t-1})^\textrm{T}e({{x}'}))}
\end{split}
\end{equation}
Here $h_\theta(x_{1:t-1})$ is the neural model's context representation, and $e(x)$ is the embedding of $x$.

For a given sequence $x$ AE language models, e.g., BERT creates a corrupt sequence $\hat{x}$ where 15\% random tokens are replaced with an artificial token [SEP]. Then the model is trained to recreate $\bar{x}$ from $\hat{x}$ such as in Equation \ref{eq:AE_models}:

\begin{equation}
\begin{split}
\underset{\theta}{\mathrm{max}}\quad log\ p_\theta(x'|\hat{x})\approx \sum_{t=1}^{T} m_t log\ p_\theta(x_t|\hat{x})
= \sum_{t=1}^{T} m_t log\frac{\mathrm{exp}(H_\theta(\hat{x})_t^\textrm{T}e(x_t))}{\sum_{{x}'}\mathrm{exp}(H_\theta(\hat{x})_t^\textrm{T}e({{x}'}))}
\label{eq:AE_models}
\end{split}
\end{equation}

Both of these models have the pros and cons. For example, BERT assumes all tokens are independent and make noise by replacing input with an artificial symbol like [MASk]; AR models don't have contextual access for both sides of the tokens. Addressing all of these, XLNet built a permutation language model that pre-trains the data like an AE model by combining the advantages of AR models. The objective can be expressed using the Equations \ref{eq:PLM}.

\begin{equation}
    \label{eq:PLM}
    \underset{\theta} {\mathrm{max}}\quad \mathbb{E}_{z\sim z_T} \left [ \sum_{t=1}^{T} log\ p_\theta(x_t|\mathbf{x}_{<t}) \right ]
\end{equation}

Here $x$ is the text sequence, and $z$ represents the sample factorization. $p_\theta(x)$ is the maximum likelihood of sequence $x$ in the accordance of $z$.

XLNet uses two-stream self-attention using two different types of hidden representation $h_\theta(x_{z_{\le t}})$ and $g_\theta(x_{z_{<t}}, z_t)$.

\begin{figure*}
    \centering
    \includegraphics[width=.8\linewidth]
    {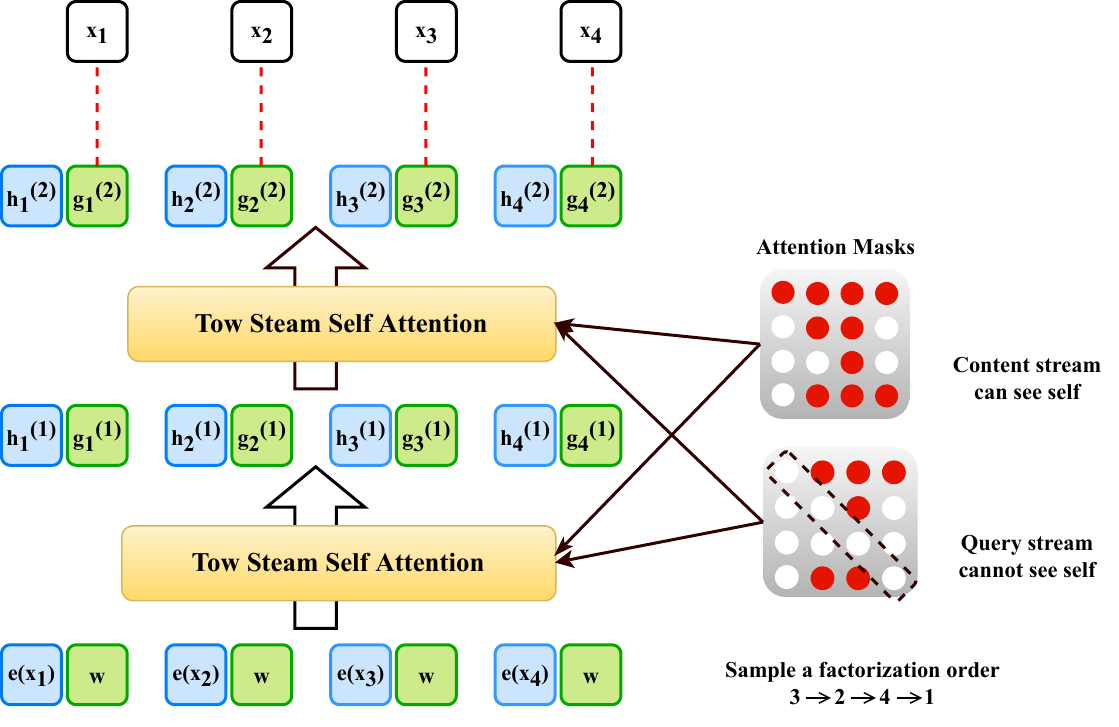}
    \caption{Permutation Language Model for predicting token $x_3$ for a given factorization order.}
    \label{fig:PLM}
\end{figure*}

Figure \ref{fig:PLM} is the illustration of the overview of a Permutation Language Model for two-stream self-attention following parameters are reformed schematically.

$g_{z_t}^{(m)}$ Attention(Q = $g_{z_t}^{(m-1)}$), KV = $h_{z_{<t}}^{(m-1)}$; $\theta$)

$h_{z_t}^{(m)}$ Attention(Q = $h_{z_t}^{(m-1)}$), KV = $g_{z_{\le t}}^{(m-1)}$; $\theta$)

Initially, $h_{i}^{(0)} = e(x_{i}) $ and $g_{i}^{(0)} = w$ and Q, K, and V are, respectively Query, Key, and Value.

\subsubsection{HowSumm Dataset}
HowSumm \cite{boni2021howsumm} is a large-scale dataset based on the wikiHow articles. The dataset has two parts: steps and methods. The observations of this dataset contain information such as the article's URL, title, a target summary for summarization tasks, method name, steps, the source, and a list of categories.
In this research, we have focused on the step part, which contains 11,121 observations and almost 6000+ unique labels for multilabel classification tasks. The challenge with this dataset is that there are lots of labels for which enough observations are not present. Therefore, before the utilization of the dataset, it was necessary to prepare it for deep learning models.

\subsection{Proposed Methodology}
The proposed methodology of this research has been described here. We can divide the methodology into three parts: data preparation, fine-tuning the XLNet model, and evaluating the model. Figure \ref{fig:sysArc} illustrates the working flow of our research work.

\subsubsection{Data Preparation and Preprocessing Phase}

This phase of our methodology can be divided into more parts: label filtering, text preprocessing, tokenization, and label encoding. These steps are described below:

\begin{table}[]
\centering
\renewcommand{\arraystretch}{1.7}
\caption{Filtered Labels after applying the data preparation algorithms. All of these 67 labels are able to score more than 1, according to the algorithm. Only these few labels have more than 500 observations in the HowSumm dataset.}
\label{tab:label lis}
\begin{tabular}{|lll|}
\hline
College University \&   Postgraduate & Dog Training  & Birds \\
Dog Behavior & Youth & Work World \\
Endocrine System Health & Recipes & Sports and Fitness \\
Personal Care and Style & Cars & Outdoor Recreation \\
Arts and Entertainment & Travel &  Pets and Animals \\
First-Aid \& Emergency Health Care & Crafts & Featured Articles \\
 Individual Sports& Cats & Home Maintenance \\
Musculoskeletal System Health & Government & Youth Dating \\
Psychological Health & Cosmetics & Emotions \& Feelings \\
Banks and Financial Institutions & Fashion & Babies and Infants  \\
Infectious Diseases & Subjects & Gardening \\
 Nutrition and Food Health & Women’s Health & Horses \\
Cars \& Other Vehicles & Legal Matters & Dating \\
Communication Skills & Skin Care &  Music \\
 Digestive System Health  & Dogs & Development Stages \\
Education and Communications & Relationships & Home and Garden \\
Endocrine System Health & Sleep Health & Parenting\\
Computers and Electronics & Hair Care &  Finance \& Business \\
Psychological Disorders & Family Life & Hobbies and Crafts \\
Integumentary System Health & Cleaning & Studying \\
Coping with Illness & Anxiety & Health  \\
Food and Entertaining & Housekeeping &  Work World \\
Cardiovascular System Health & Business & Sports and Fitness \\
Outdoor Recreation & Team Sports &  
\\
\hline
\end{tabular}
\end{table}

\begin{itemize}
    \item  \textbf{Label Filtering: }We preprocess the text using different preprocessing techniques. Preparing our label presents several challenges. The dataset contains more than 6,000 labels. However, most of these labels don't have enough observations, which tends to imbalance the dataset and affect the classifier's performance.
Therefore, we apply the proposed Algorithm \ref{alg_Data_prepa} and prepare our data for this research. In this dataset, the least number of observations for any single label is only three. To tackle this data imbalance problem, we have chosen a methodical approach. We are removing the labels with a smaller number of observations using a specific algorithm. Our aim is to retain only those labels that have at least 500 observations, a strategy that we believe will effectively address the imbalance. This is how we have ensured that each label used in this paper has a sufficient amount of observations for preventing the label imbalance problem.
This algorithm takes a dataset with text and labels and then returns a dataset after filtering labels. 
In Line 1, we are getting the List of selected labels using the procedure $Label\_Selection$, which is implemented in Lines 12-26. Here, we have identified a selection score for each label using a threshold value, which is 500 for this research, and returned a list of selected labels that scored more than or equal to 1. The method uses a priority queue named $ List_{Selected 
label}$ and storing the filtered labels. If we increase the threshold value, we need to get more labels from the dataset for multiclass classification. Moreover, by decreasing the threshold value we face, the class creates a class imbalance problem in our experimental results.
Line 21 calculates the selection score for a particular label.
Table \ref{tab:label lis} showed the List of selected 67 labels using the method $Label\_Selection$.
In Lines 2-11, we relabeled our texts using these chosen labels. Here, a while loop is used to iter over the text data. In each iteration, the algorithm checks whether a label for the text is present in $ List_{Selected 
label}$ or not. If it is present, the algorithm keeps the label for that text. Otherwise, the label is dropped. In a similar manner, all the labels related to that text are checked.
After filtering labels, our text is labeled with more than one label, which enables the multiclass classification problem in this research.

\item \textbf{Text Tokenization:} We preprocessed the raw text, which will be discussed later in the paper with proper examples. After that, we sent our text data to the XLNet and BERT tokenizer, as direct text cannot be absorbable by any model. Both tokenizers tokenize the data and some special token \texttt{[CLS]}, \texttt{[PAD]} and prepare the input sequence for corresponding classifiers. We get three types of sequences from the tokenizer: input sequence, attention mask, and segment IDs.

\item \textbf{Label encoding:} For encoding our labels, we have selected a binary approach. For each text, we have created a binary list with the length of the total selected levels. The index of this array is the representation of all the labels' numbers. For a text, a label is assigned to that data; then, the array value will be one based on the label's number; otherwise, the value is 0.

\end{itemize}

The workflow of this part is mentioned in the "Data Preparation and Preprocessing" phase of Figure \ref{fig:sysArc}.

\begin{algorithm}
\algorithmicrequire { $Data[Text, Labels]$}\\
\algorithmicensure { $Data[Text, Labels_{Selected}]$} 
 \begin{algorithmic}[1]

\State $List_{selectedLabel} \gets Label\_Selection(Data[Text, Labels]
)$
\State $i \gets 0$ 
\While{$i < Data[Labels].length$}
    \For{$each\ x_{label}\ in\ Data[Labels][i]$}
    \If{$x_{label}\ in\ List_{selectedLabel}$}
    \State $Keep\ x_{label}$
    \Else
    \State $Drop\ x_{label}$
    \EndIf
    \EndFor
\EndWhile
\Procedure{$Label\_Selection$} {$Data[Text, Labels]$}
\State $List_{selectedLabel} \gets Priority_Queue$
\State $Threshold \gets 500$
\State $j \gets 0$ 
    \While{$i < Data[Labels].length$}
        \State $List_{uniqueLabel} \gets unique\ in\ Data[Labels][j]$ 
    \EndWhile
    \For{$each\ y_{label}\ in\ List_{uniqueLabel}$}
    \State $total_{y_{label}} \gets Compute\ the\ total\ no\ of\ text\ for\ y_{label}$
    \State $Selection\_Score \gets \frac {total_{y_{label}}}{Threshold}$
    \If{$Selection\_Score \geq 1$}
    \State $List_{uniqueLabel}.add(y_{label})$
    \EndIf
    \EndFor
    \State $return List_{uniqueLabel}$
\EndProcedure
 \caption{Algorithm for Data Preparation}\label{alg_Data_prepa}

\end{algorithmic}
\end{algorithm}

\subsubsection{Fine Tuning the Model}

The "Model Training" phase of Figure \ref{fig:sysArc} represents this part of our methodology. 
Here, we fine-tuned the BERT and XLNet models for multi-label classification. Both models consist of two input layers. One is for the input sequence, one for the attention mask, and the other for the segment IDs. Through the input layer, the inputs are passed to the XLNet layer for the XLNet classifier. In  BERT architecture, the next layer after the input layers is the BERT layer. A fully connected layer is also used after that, and finally, the activation layer is used to predict the labels. We use the 'Sigmoid' activation function here in both models. 

We trained the models with tokenized text and binarized labels. By manipulating the test data, we attained two different evaluation measures: binary cross-entropy loss and binary accuracy. The binary accuracy indicates the correct prediction of positive and negative values among all the predictions. Equation \ref{bin acc} is used to calculate binary accuracy.

\begin{equation}
\label{bin acc}
    Accuracy=\frac{TN+TP}{TN+TP+FN+FP}
\end{equation}

Here, $TN$ is the True Positive, which means the total correct prediction of true labels for each text. Then comes the $FP$ False Positive, which is the number of wrong predictions of true labels. $TN$, the True Negative, and $FN$, the False Negative, are the number of consecutively true and false prophecies of the 0 class.

Averaging the log of the probability of right predictions, the Binary Cross Entropy Loss works. 
Based on that, we backpropagate the models and detect the necessary weights for the prediction of unseen data. It identifies the probability distribution of the predictions and ground truth labels. Equation \ref{bin loss} used to calculate Binary Cross Entropy Loss.

\begin{equation}
\label{bin loss}
   \mathbb{L }_{Binary}= -\frac{1}{N}\sum_{i=1}^{N} y_i.log(p(y_i))+(1-y_i).log(1-p(y_i))
\end{equation}
Here $p(y_i)$ and $1-p(y_i)$ are the probability of zero and ones in the predicted labels.

Hyperparameters play a significant role in model fine-tuning, as they provide control over the learning process and the parameter values of a model. Therefore, we tuned the hyperparameters and compared the accuracy of the modifications for both XLNet and BERT models. Doing so helped us to perceive the best values for our proposed models in this research.

\subsubsection{Model Validation}

Finally, the trained XLNet and BERT models are utilized for validation. These trained models consist of necessary weights for the predictions.
We sent an unseen to both models. Then, we processed and tokenized the text to send it to the trained model. 

\begin{figure*}
    \centering
    \includegraphics[width=.8\linewidth]
    {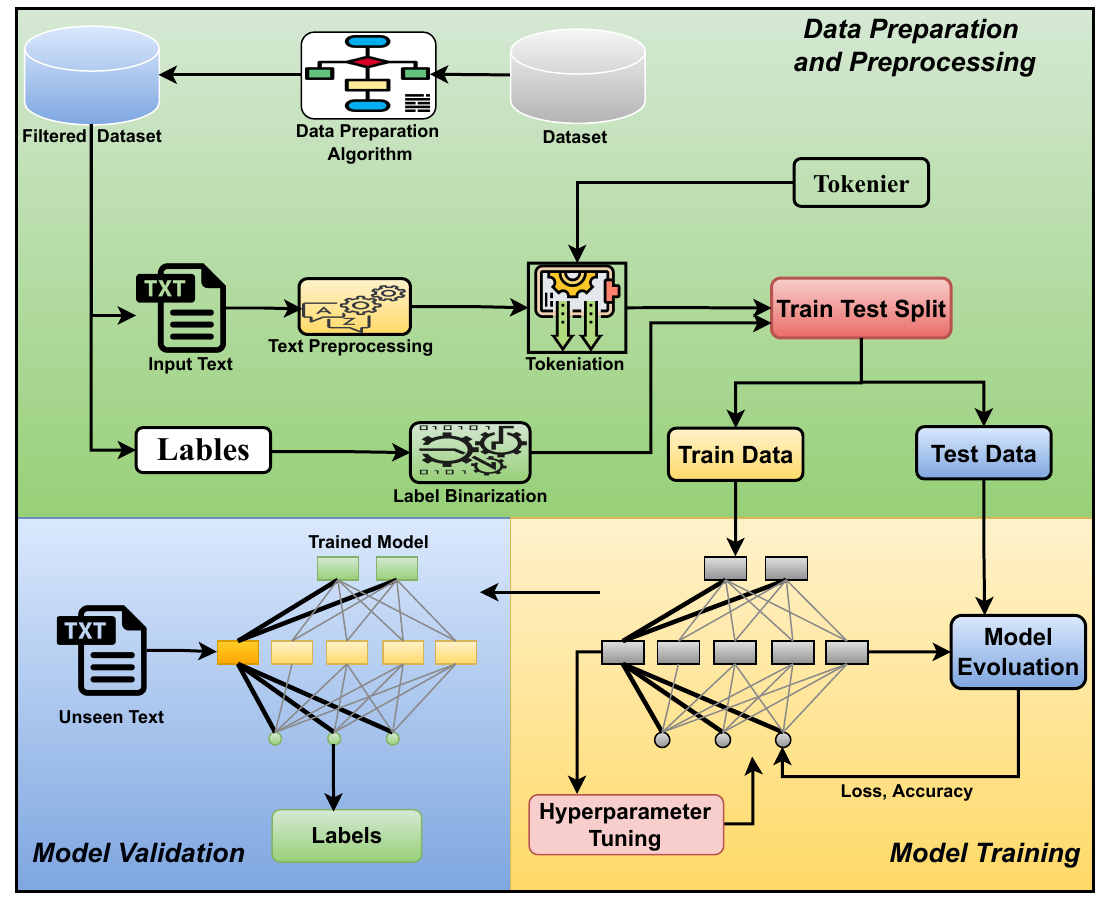}
    \caption{The system architecture of our proposed methodology. It indicates the system's workflow and is divided into three parts- Data Preparation and Preprocessing (Workflow given inside the green box of the figure), Model Training (The yellow box of the figure) and Validation (The blue box of the figure).}
    \label{fig:sysArc}
\end{figure*}

Finally, the model worked to get the prediction of multiple labels from the given input. This part is illustrated in the model validation phase of Figure \ref{fig:sysArc}.
Alongside binary accuracy and Binary Cross Entropy Loss, we have employed other metrics such as Macro Average F1 Score, Micro average F1 Score, Macro Average Precision, and Macro Average Recall. To calculate the Macro Average F1 Score and Micro average F1 Score, we rely on the precise Equations \ref {eq:maf1} and \ref {eq:maf2}.

\begin{equation}
    \label{eq:maf1}
Macro\ Average\ F1\ Score=\frac{2\times P_{MA}\times R_{MA}}{P_{MA}+ R_{MA}}
\end{equation}

\begin{equation}
    \label{eq:maf2}
Micro\ Average\ F1\ Score=\frac{\overline{TP}}{\overline{TP}+\frac{1}{2}(\overline{FP}+\overline{FN})}
\end{equation}

Here, $P_{MA}$ and $R_{MA}$ are the macro average precision and recall. Macro Average Precision and Recall, calculated by averaging each class's precision and recall value. $\overline{TP}$, $\overline{FP}$, and $\overline{FN}$ are the net True Positive, False Positive, and False Negative counted from the confusion matrix of each class.

\section{Experimental Setup}
\subsection{Experimental Environment}

We use Google Colab (https://colab.research.google.com/) to train our image analysis or classification model, which requires high computing powers that GPU (Graphics Processing Unit) can offer. However, GPU installation is costly and needs extra hardware to support the computation. Google Colab provides us with a high-end GPU on the cloud, along with all the essential packages for the training process. We do not have to install packages or concerns about storage space \cite{bisong2019google}. The specs of Google Colab include NVIDIA K80 GPU, 12 GB of GPU memory, Up to 2.91 teraflops double-precision rendition, and 358 GB of disk space. These specs create a robust computation environment for training Deep Learning models.

\subsection{Hyperparameter Tuning}
It is appreciated that the hyperparameters of that model be tuned to bring out the best performance of a deep learning model. We train the model with five different Learning rates ( 1e-04, 2e-04, 3e-04, 4e-04, and 5e-04) and two different maximum lengths of the text sequence (484 and 512). We use the AdamW \cite{kingma2014adam} optimizer in this work. The suitable parameters for this research have been enlisted in Table \ref{tab:hyper}. We considered the following hyperparameters for our models:
\begin{itemize}
    \item \textbf{Learning Rate: }A significant hyperparameter that controls how a model will adjust the updates for parameters. BERT and XLNet provide the best performance for learning rates 5e-04 and 4e-04, respectively.
    
\item \textbf{Batch Size: } It indicates the number of samples the model considers before updating the parameters during the model propagation. We consider only one batch size, which is 48 here for both models.

\item \textbf{Maximum Length: }Models usually works with fixed length sequence. The maximum length determines that for a model. The maximum length for the sequence we consider here is 512. BERT can take a maximum 512-length sequence. XLNet doesn't have any limitations for text sequence. It can take any length of the sequence. Still, we use two different maximum lengths, 484 and 512, for both models so that we can create a comparison among transformer models. Both of these models performed their best for the maximum length of 512.

\item \textbf{Optimizer: }It is the algorithm or function that updates the model attributes to reduce the loss and improve the performance. Here we use the AdamW optimizer for both models. It is a stochastic gradient descent method that uses both first and second-order moments.
\end{itemize}

\begin{table}[!htp]
\centering
\renewcommand{\arraystretch}{1.7}
\caption{Hyperparameters for the proposed models. For these values, the models provide the highest accuracy here. (The table columns are the Hyperparameter name and most significant Hyperparamters value for XLNet and BERT)}
\label{tab:hyper}
\begin{tabular}{|l|c|c|}
\hline
\textbf{Hyperparameters} & \textbf{XLNet} & \textbf{BERT} \\ \hline
Learning Rate & 4e-04 & 5e-04 \\ \hline
Batch size & 48 & 48 \\ \hline
Mximum Length & 512 & 512 \\ \hline
Optimizer & AdamW & AdamW \\ \hline
\end{tabular}
\end{table}

\subsection{Data Preprocessing}

\begin{table*}[!htp]
\centering
\renewcommand{\arraystretch}{1.07}
\caption{Dataset state after applying our data preparation algorithm.}
\label{tab:data}
\begin{tabular}{|l|c|}
\hline
\multicolumn{1}{|c|}{\textbf{Text}} &
  \textbf{Selected Label} \\ \hline
\begin{tabular}[c]{@{}l@{}}Type the author's last name first,\\ followed by a comma. Then type the\\ author's first initial. Add their\\ middle initial, if given. Type a space \\ after the period, then type  the date\\ of publication in parentheses. \\ Include the year first, followed by a\\ comma, then the month and day \\ (if provided). Place a period after the\\ closing parentheses. Example: \\ Will, G. F. (2004, July 5). If there\\ are multiple authors, separate their \\ names with commas. Use an ampersand \\(\&) before the last author's \\name.\\ \end{tabular} &
  \begin{tabular}[c]{@{}c@{}}Education and Communications, \\ College University and Postgraduate\end{tabular} \\ \hline 
\begin{tabular}[c]{@{}l@{}}Don’t be in a rush to get up after a\\ fainting spell. Your body and mind\\ need time to recover. You should stay\\ in your current position on the \\ground for at least 10-15 minutes. If\\ you get up too soon you risk \\triggering another episode.\end{tabular} &
  \begin{tabular}[c]{@{}c@{}}Health, \\ Cardiovascular System Health\end{tabular} \\ \hline
\begin{tabular}[c]{@{}l@{}}Keep track of how many hands it takes\\ to go from the bottom of the sun\\ to the horizon. The number of hands\\ it takes is the number of daylight\\ hours remaining, or the hours left\\ until sunset. For example, if you count\\ 5 hands, then there are 5 hours \\remaining in the day or 5 hours until \\sunset.\end{tabular} &
  \begin{tabular}[c]{@{}c@{}}Home and Garden, \\ Featured Articles'\end{tabular}\\ \hline 
\end{tabular}
\end{table*}

Before tokenizing our text and applying the data preparation algorithm, we preprocessed the data. Table \ref{tab:data} shows some samples of data after applying the Algorithm \ref{alg_Data_prepa}. We present four samples in this table, and multiple labels are given for each text.
In these data, we have used the following preprocessing methods:

\begin{itemize}
    \item Raw text has lots of special characters (.,\%\@\&\*\# etc). First, we remove all the special characters from our data. These characters are completely unnecessary and have no contribution to the model's performance.
    \item Another component of any text that has no relevance to the model's performance is the stop word. Therefore, we eliminated all the stop words from our text. 
    \item For more efficiency, we have lemmatized our text and used the lemmas of different complex words.
    \item Raw text is the mixture of both capital and case letters. Finally, we convert all the text into lowercase. It will be beneficial for improving the performance of both classifiers.
\end{itemize}

We have mentioned our raw and preprocessed below for the visualization of the changes in data \cite{ahmedasonam2021}.

\begin{itemize}
    \item \textbf{Raw Data}\\ Type the author's last name first, followed by a comma. Then, type the author's first initial. Add their middle initial, if given. Type a space after the period, then type the date of publication in parentheses. Include the year first, followed by a comma, then the month and day (if provided). Place a period after the closing parentheses. Example: Will, G. F. (2004, July 5). If there are multiple authors, separate their names with commas. Use an ampersand (\&) before the last author's name.
    \item \textbf{Prepossessed Data}\\ type author last name first follow comma then type author first initial add middle initial give type space period type date publication parentheses include year first follow comma month day provide place period closing parentheses example will g f 2004 July five if multiple author separate name comma use ampersand last author-name
\end{itemize}

\section{Experimental Results and its Comparison}

\subsection{Model Performance}
We observed the model's performances in different measures during the model training and applied the unseen validation data. For hyperparameter tuning, we determine the accuracy of different hyperparameters. The estimation given in Table \ref{tab:hyper} provides the highest performance. 
For both BERT and XLNet, we determine accuracy scores for different hyperparameter values. Table \ref{tab:accHyper} shows different accuracy values for different hyperparameters. The highest accuracy for XLNet is 97.30\%, and for the BERT model, it is 97.17\%. In the Table, both of these values are marked as bold.

\begin{table*}[]
\centering
\renewcommand{\arraystretch}{1.7}
\caption{Different Binary Accuracy Score of the proposed XLNet and BERT models for different hyperparameter values. (The table's columns are Learning Rate, Maximum Length, XLNet Accuray and BERT Accuracy )}

\label{tab:accHyper}
\begin{tabular}{|l|c|c|c|}
\hline
\multicolumn{1}{|c|}{Learning rate} & Max Length & \begin{tabular}[c]{@{}c@{}}XLNet\\ Accuracy(\%)\end{tabular} & \begin{tabular}[c]{@{}c@{}}BERT\\ Accuracy(\%)\end{tabular} \\ \hline
\multirow{2}{*}{1e-04} & 484 & 96.50 & 96.03 \\ \cline{2-4} 
 & 512 & 96.69 & 96.22 \\ \hline
\multirow{2}{*}{2e-04} & 484 & 95.90 & 96.91 \\ \cline{2-4} 
 & 512 & 96.01 & 96.22 \\ \hline
\multirow{2}{*}{3e-04} & 484 & 97.06 & 95.60 \\ \cline{2-4} 
 & 512 & 97.13 & 97.00 \\ \hline
\multirow{2}{*}{4e-04} & 484 & 96.98 & 95.54 \\ \cline{2-4} 
 & 512 & \textbf{97.30} & 96.78 \\ \hline
\multirow{2}{*}{5e-04} & 484 & 97.00 & 96.03 \\ \cline{2-4} 
 & 512 & 97.21 & \textbf{97.17} \\ \hline
\end{tabular}
\end{table*}

We track the model's loss and accuracy over 40 epochs. This helps to understand how well the model is built over epochs. Here, we used 'Binary Accuracy' and 'Binary Cross Entropy Loss' for the loss. Binary accuracy calculates the percentages of correct classes among all predictions in binarized labels. 

Our decision to plot the Training and Testing accuracy of our models over 40 epochs has proven to be a valuable strategy. This visualization of the learning process offers a clear picture of how the model has evolved. Most importantly, it allows us to detect any underfitting or overfitting, ensuring the model's performance is optimal.

\begin{figure}
    \centering
    \includegraphics[width=.7\linewidth]{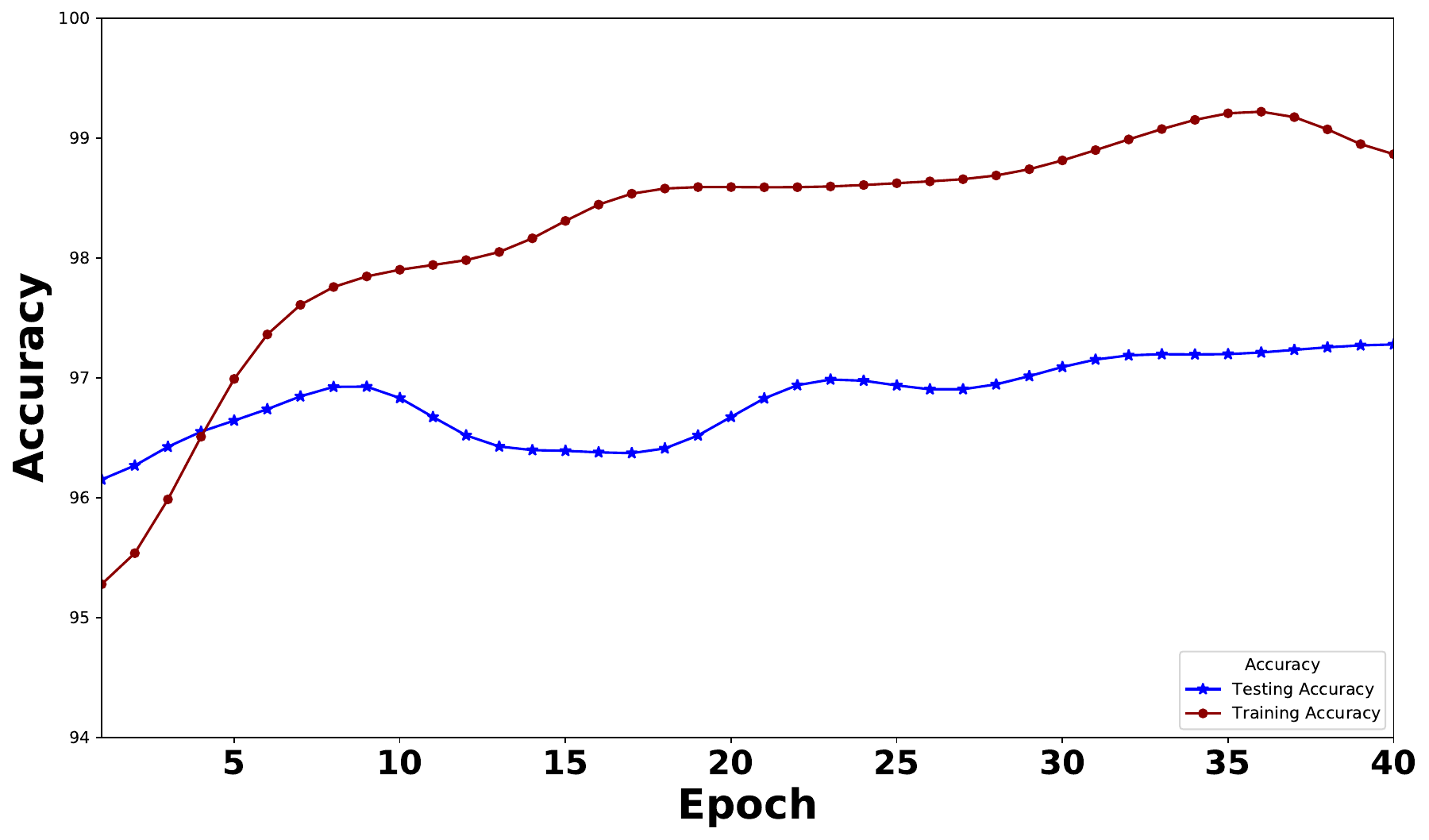}
    \caption{Training accuracy and Testing accuracy of the XLNet model over epochs. (The curve is smoothened using the Gaussian filter. )}
    \label{fig:acc graph}
\end{figure}

\begin{figure}
    \centering
    \includegraphics[width=.7\linewidth]{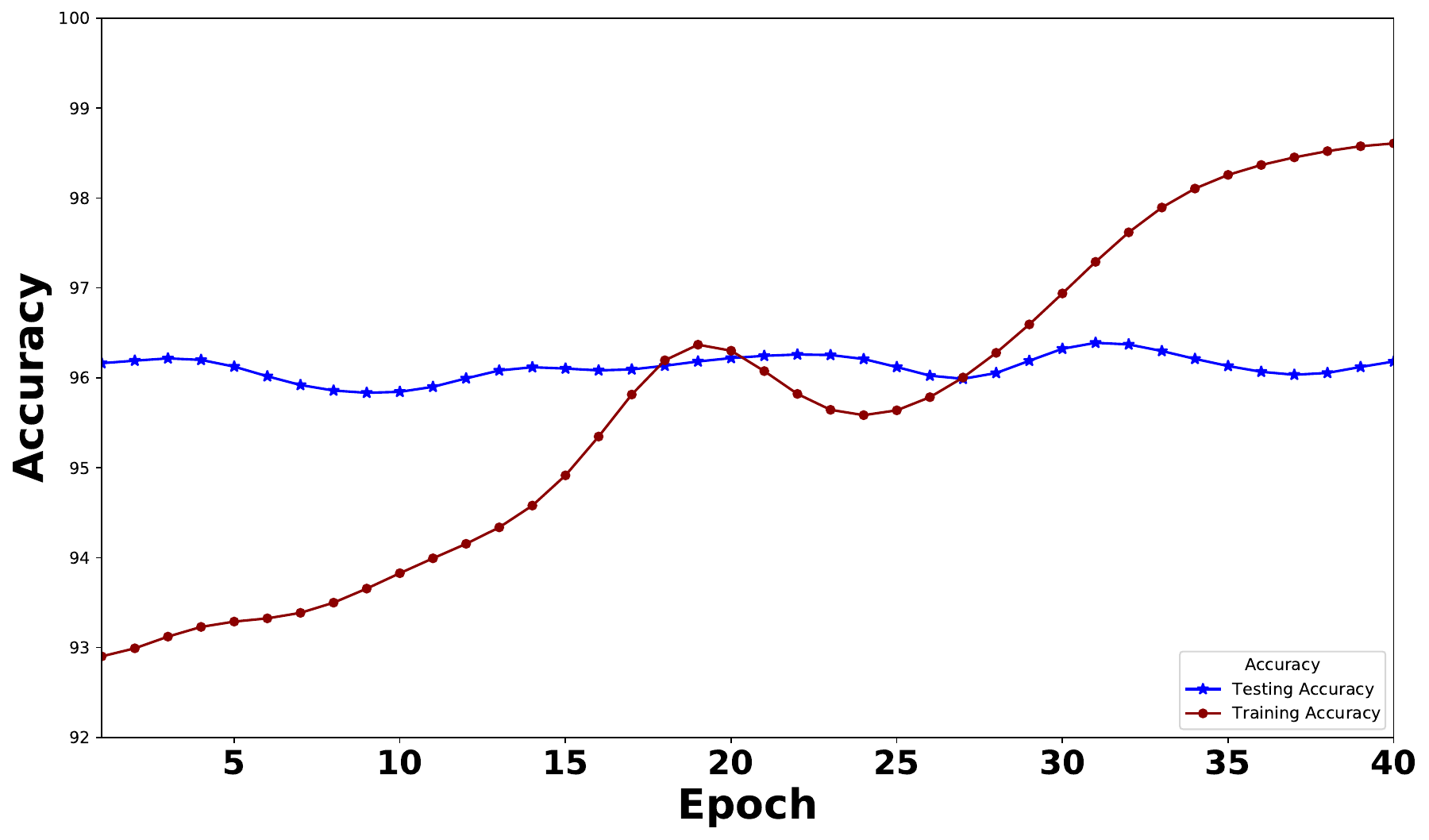}
    \caption{Training accuracy and Testing accuracy of the BERT model over epochs. (The curve is smoothened using the Gaussian filter. )}
    \label{fig:acc graph bert}
\end{figure}

Figure \ref{fig:acc graph} presents a visual representation of the Training and Testing accuracy of the XLNet model over 40 epochs. The blue curve represents the testing accuracy, while the red curve depicts the training accuracy. The highest training accuracy and testing accuracy, 99.47\% and 97.31\% respectively, serve as clear benchmarks for the model's performance.
In a similar way, the BERT model's Training and Testing accuracy is given in Figure \ref{fig:acc graph bert}. Here, the highest training accuracy is 98.76\% and 97.14\% over 40 epochs. After observing these two curves, we can state that our model is neither underfitted nor overfitted.

We have also presented the losses over epochs for our models. The training loss is a key tool for detecting how the training data has been adjusted in the model, and it is used to gauge the model's performance in the validation dataset. Importantly, these losses can also help us identify potential underfitting and overfitting in our models, making us aware of their limitations.

\begin{figure}
    \centering
    \includegraphics[width=.7\linewidth]{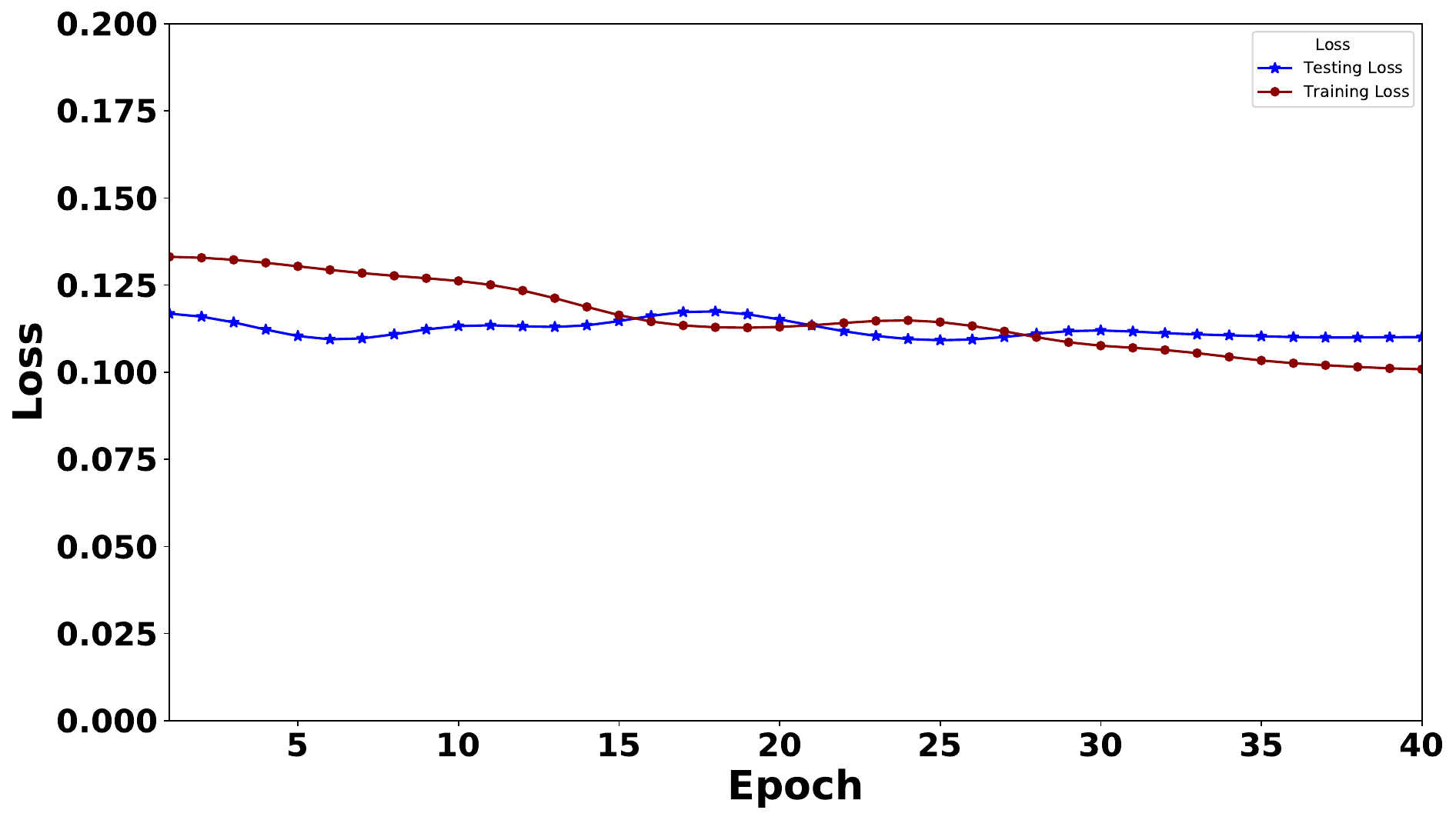}
    \caption{Training loss and Testing loss of the XLNet model over epochs. (The curve is smoothened using the Gaussian filter. )}
    \label{fig:loss graph}
\end{figure}

Figure \ref {fig:loss graph} provides a clear view of the training and testing losses for XLNet. The blue and red curves represent the testing and training accuracy over epochs, respectively. The lowest training loss is .1000, and it is .1052 for testing. These losses are crucial indicators of the model's performance, providing valuable insights into its accuracy and fit. 

\begin{figure}
    \centering
    \includegraphics[width=.7\linewidth]{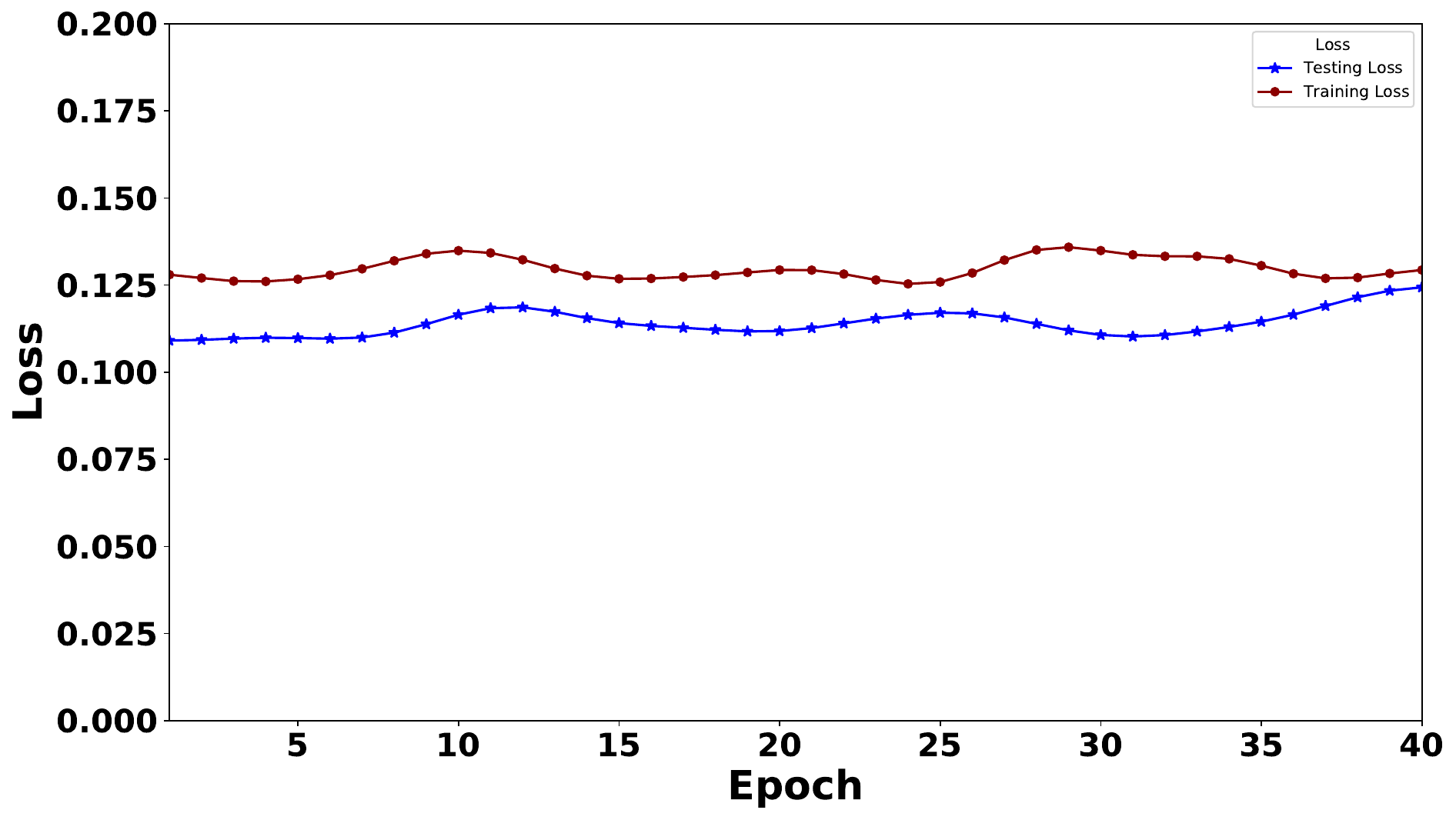}
    \caption{Training loss and Testing loss of the BERT model over epochs. (The curve is smoothened using the Gaussian filter. )}
    \label{fig:loss graph bert}
\end{figure}

Similarly, we have plotted the losses for the BERT model in Figure \ref{fig:loss graph bert}. The BERT model demonstrated a training loss of .1000, and a testing loss of 0.1103. These values, along with the curves of the models' losses, indicate that the training data have been effectively fitted to our models. Furthermore, their performance in the validation data is commendable, providing a strong basis for comparison with XLNet.

For smoothing the curve in both Figure \ref{fig:acc graph}, Figure \ref{fig:acc graph bert}, Figure \ref{fig:loss graph} and Figure \ref{fig:loss graph bert}, we use the Gaussian filter.

\begin{figure}
    \centering
    \includegraphics[width=1\linewidth]{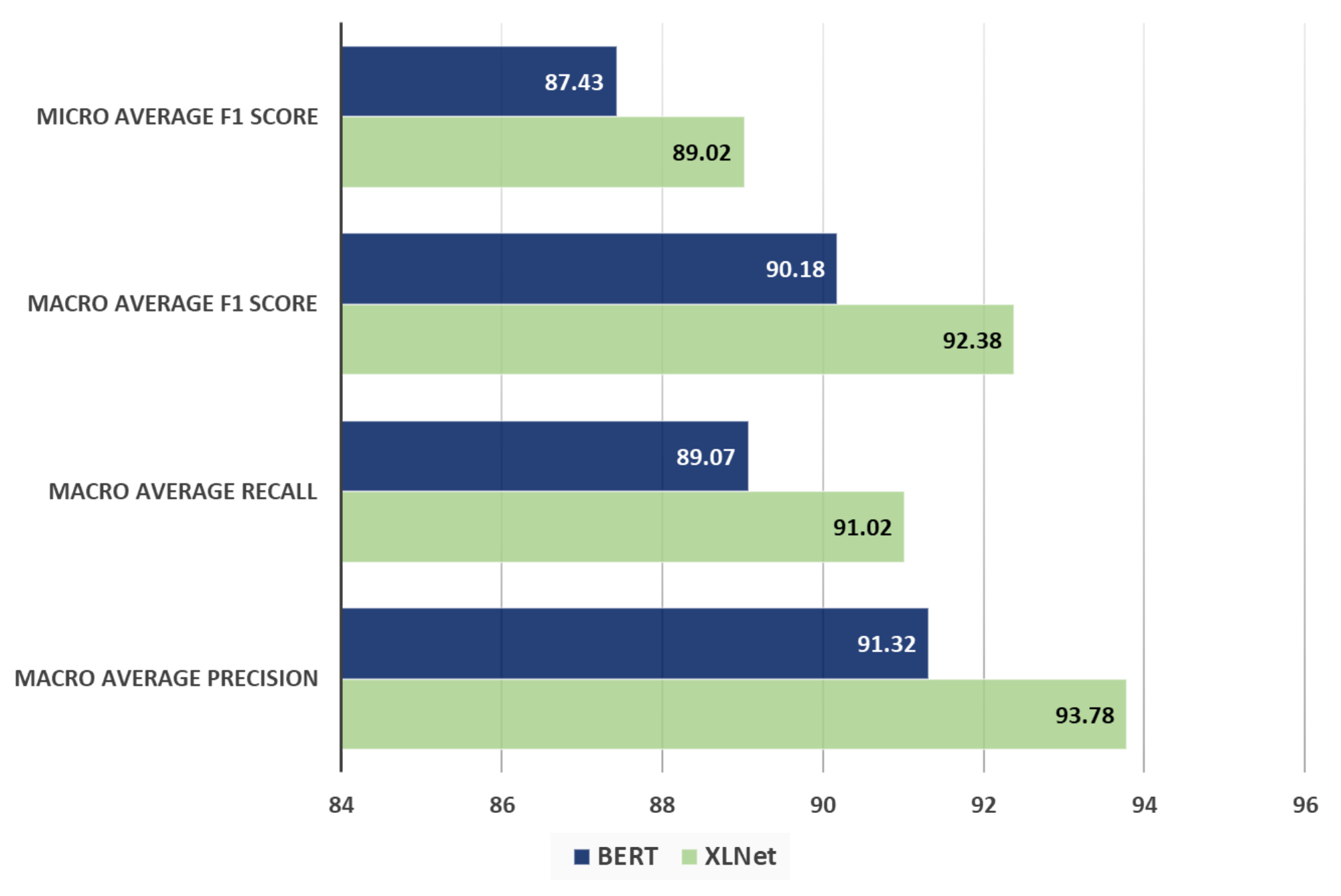}
    \caption{Evaluation Metrics for HowSumm Dataset}
    \label{fig:evaluation metric}
\end{figure}

Our work goes beyond mere accuracy. We have meticulously determined a range of evaluation metrics, each playing a crucial role in our model evaluation. In figure \ref{fig:evaluation metric}, we have sketched a bar chart where bars represent Macro Average Precision, Recall and F1 Score, and Micro Average F1 Score for our selected classifier XLNet and BERT.

\subsection{Comparison with Other Models and Dataset}

To compare the proposed method, we implemented other deep learning models: BERT, ELECTRA, Roberta, DistilBert, DeBERTa, GPT-4 and LSTM. Among all of these models, XLNet outperforms others. Here, we use binary accuracy and the Macro and micro average F1 scores.
For the transformer models, we use their corresponding tokenizer to tokenize the sequence and then applied to the model. 

For comparison, at first, we implemented RoBERTa \cite{liu2019roberta}. It is a variation of BERT where the pretraining is performed in a larger dataset and after removing the Next Sentence Prediction(NSP) task from the pretraining phase. The authors claimed that removing NSP can help to improve performance.
In our data, Roberta provided 96.90\% accuracy, the Macro average F1 score was 89.01\%, and the micro average score was 87.98\%. 

The next transformer architecture is ELECTRA \cite{clark2020electra}. It is very similar to BERT; however, unlike BERT, it used a generator and discriminator during the pretraining phase. The generator performed like the Mask Language Model, and the discriminator aims to discover the original and predicted Tokens. Here, ELECTRA has shown 95.98\% accuracy, 87.56\% Macro average F1 score, and 86\% micro average score.

Another transformer-based model, DistilBert \cite{sanh2019distilbert}, performed with 97.02\% accuracy, 90.2\% Macro average F1 score, and 87.43\% micro average F1 Score. Compared to BERT, it is cheap, small, and fast. It uses fewer parameter values than BERT. 

We also applied DeBERTa\cite{hedeberta} and GPT-4 to instructional text. These transformer models were introduced recently and performed well on different tasks. Here, we can see that these two models performed very closely to XLNet.  Even DeBerta provided the highest accuracy but not the highest F1 scores. The accuracy of DeBERTa and GPT-4 are 97.32\% and 9.20\%, macro average F1 Scores are 90.34\% and 91\%, and micro average F1 Scores are 88.01\% and 89.56\%.

We also applied the LSTM(Long Short Term Memory), an autoregressive model. LSTM can handle shorter sequences than transformers and uses different text vectorization techniques. First, we determined the word embeddings of our text and then applied them to LSTM.
The accuracy, macro average F1 score, and micro average F1 score are 95.8\%, 85.09\%, and 82.09.

\begin{figure*}
    \centering
    \includegraphics[width=1\linewidth]{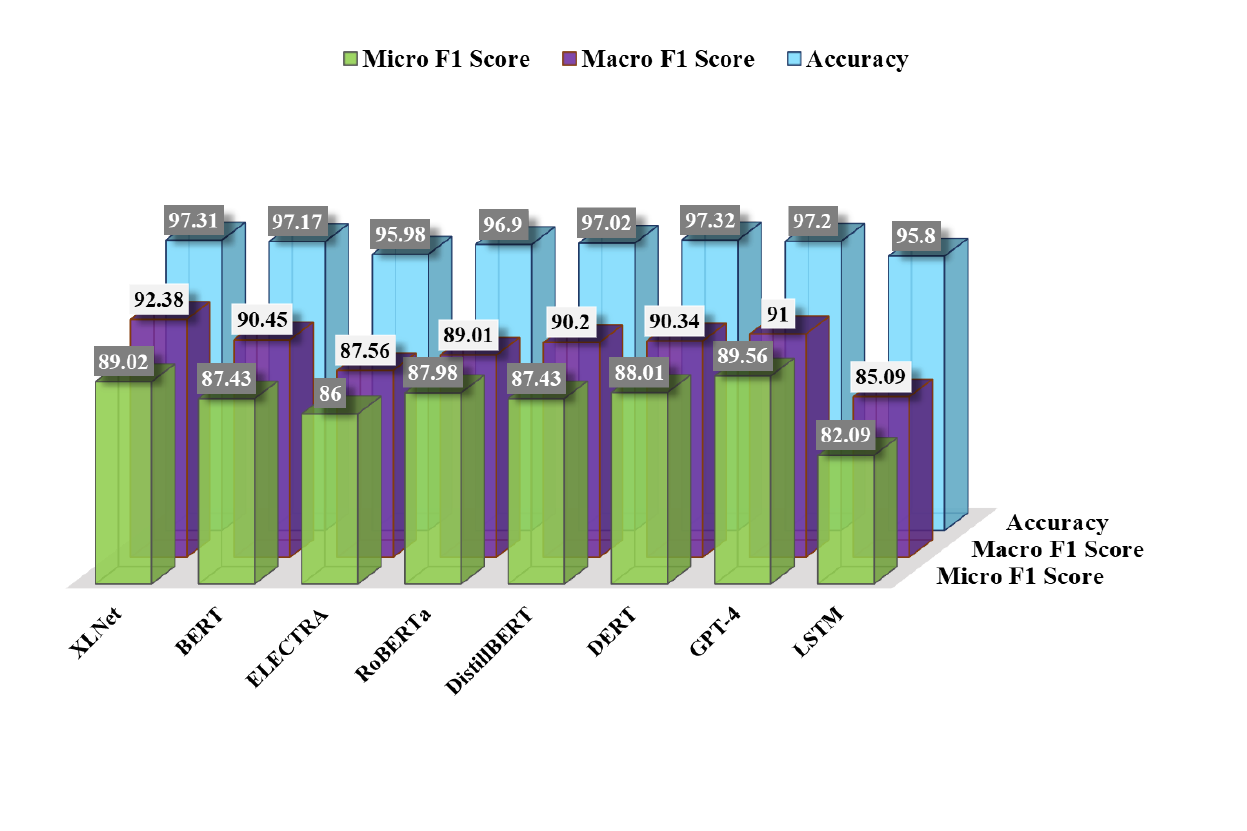}
    \caption{Accuracy, Macro and Micro F1 Score of different deep learning models.}
    \label{fig:model comp}
\end{figure*}

Figure \ref{fig:model comp} represents the summary of binary accuracy, Macro and micro F1 score of the data for different deep learning models. We can observe that our proposed models, XLNet and BERT, outperform all other models.

We also have applied our methodology to other datasets of multilabel classification named 'arXiv Paper Abstracts' dataset(https://www.kaggle.com/datasets/spsayakpaul/arxiv-paper-abstracts), Twitter Financial News (https://www.kaggle.com/datasets/sulphatet/twitter-financial-news), Toxic Comment Classification (https://www.kaggle.com/c/jigsaw-toxic-comment-classification-challenge/data). Table \ref{tab:paper dataset} represents different evaluation metrics values for our classifier on these datasets. Our methodology performed well for these datasets, which is clearely visualized from Table \ref{tab:paper dataset}.

\begin{table}[]
\centering
\caption{Evaluation Metrics of three different multi-lable datasets for BERT and XLNet. }
\label{tab:paper dataset}
\begin{tabular}{|l|cc|cc|cc|}
\hline
\multicolumn{1}{|c|}{\textbf{Dataset$\longrightarrow$}} & \multicolumn{2}{c|}{\textbf{\begin{tabular}[c]{@{}c@{}}arXiv Paper \\ Abstracts\end{tabular}}} & \multicolumn{2}{c|}{\textbf{\begin{tabular}[c]{@{}c@{}}Twitter Financial \\ News\end{tabular}}} & \multicolumn{2}{c|}{\textbf{\begin{tabular}[c]{@{}c@{}}Toxic Comment \\ Classification\end{tabular}}} \\ \hline
\multicolumn{1}{|c|}{\textbf{Metrics $\downarrow$}} & \multicolumn{1}{c|}{XLNet} & BERT & \multicolumn{1}{c|}{XLNet} & BERT & \multicolumn{1}{c|}{XLNet} & BERT \\ \hline
Accuracy & \multicolumn{1}{c|}{99.59} & 99.46 & \multicolumn{1}{c|}{98.45} & 99.01 & \multicolumn{1}{c|}{90.32} & 91.94 \\ \hline
\begin{tabular}[c]{@{}l@{}}Macro Average \\ Precision\end{tabular} & \multicolumn{1}{c|}{98.02} & 98.11 & \multicolumn{1}{c|}{94.78} & 97.23 & \multicolumn{1}{c|}{71.34} & 79.34 \\ \hline
\begin{tabular}[c]{@{}l@{}}Macro Average \\ Recall\end{tabular} & \multicolumn{1}{c|}{96.48} & 95.76 & \multicolumn{1}{c|}{96.00} & 98.29 & \multicolumn{1}{c|}{75.31} & 81.29 \\ \hline
\begin{tabular}[c]{@{}l@{}}Macro Average \\ F1 Score\end{tabular} & \multicolumn{1}{c|}{97.24} & 96.92 & \multicolumn{1}{c|}{93.10} & 97.05 & \multicolumn{1}{c|}{72.03} & 77.33 \\ \hline
\begin{tabular}[c]{@{}l@{}}Micro Average \\ F1 Score\end{tabular} & \multicolumn{1}{c|}{93.22} & 94.89 & \multicolumn{1}{c|}{92.21} & 96.07 & \multicolumn{1}{c|}{70.45} & 77.03 \\ \hline
\end{tabular}
\end{table}

\section{Discussion}
This research introduces an impactful methodology that can be instrumental in the development of multilabel instructional text classification systems. It leverages popular deep learning models such as BERT and XLNet, marking a significant advancement in this field. 
To the best of our knowledge, this is the first attempt at multilabel instruction classification.
This research's significant contribution lies in its ability to determine the appropriate tags for different instructions. This can greatly enhance instruction-searching experiences. Moreover, it paves the way for task-oriented learning of procedural text, empowering intelligent systems to understand the categories of the instruction.

This research uses one of the latest datasets based on the wikiHow article Howsumm. Though the dataset is prepared to focus on summarization tasks, we intend to utilize the dataset for classification tasks. For that, we introduce a novel and simple algorithm that filters the labels of HowSumm data and removes the class imbalance problem from it. This algorithm provided us with 67 significant labels after scoring all the labels we used for multilabel classification. In the future, we want to share this filtered dataset publicly. This helped us to ignore the data imbalance problem from the HowSumm dataset, which helped the classifier improve its performance.
Perhaps sharing this filtered data creates new possibilities for using wikiHow articles in multilabel text classification.

For the sake of simplicity and ease of classification, we have opted for the binary label encoding technique. This technique converts the labels for a single text observation into 67 long arrays, where all the elements are either 0 or 1. As a result, we have used binary accuracy and binary cross-entropy loss as our proposed methodology. To align with this methodology, the activation function used in the models is Sigmoid.

As this is the first work on instructional text classification, we here used the most significant models. We have selected the best two transformers in our search.
The latest Large Language Model for XLNet has been employed to bring out the best performance in our data. During the model building, we traced the model's training and testing measures(Loss and Accuracy) and plotted these values in a graph. Figure \ref{fig:acc graph} and \ref{fig:loss graph} indicate these graphs where we have observed how perfect the model's accuracy and loss values are over 40 epochs.
BERT is also an outperformer; the accuracy and loss graph for BERT is given in Figure \ref{fig:acc graph bert} and \ref{fig:loss graph bert}.

They provided significant scores for the data and outperformed other transformer models such as ELECTRA, Roberta, DistilBert, and another deep learning model, Long Short-Term Memory (LSTM). We maneuver two evaluation techniques for the justification of the proposed methodology: Accuracy, Macro, and Micro Average F1 Score. 
Figure \ref{fig:model comp} shows how XLNet outperforms other models with an Accuracy of 97.30\%. The best performer, XLNet, is a generalized autoregressive transformer, which combines the facility of an autoencoding and an autoregressive model's facilities and avoids their limitations. For the comparison, we have implemented BERT, an autoencoding Large Language Model, and LSTM, an autoregressive model. XLNet outperforms both of them with higher accuracy and Macro F1 Score. Though XLNet has no sequence length limit, it still outperformed the models with fixed length 512. As the instructional texts are longer, XLNet performed better than BERT.
 
This research marks the beginning of a new era in instructional text classification. By harnessing the power of the most advanced Large Language Model, we have kept the overall process straightforward. The use of binary encoding for labels, which tends to use sparse data, has led to a remarkably high-performing model. These practical implications underscore the potential of our proposed architecture in real-world scenarios. 

\section{Conclusion and Future Work}
In our research, we aim to address a notable problem that needs more attention. Specifically, we tackle the challenge of multilabel classification for wikiHow instructions. Our approach is relatively straightforward, using a binary label encoding technique. Our proposed methodology has demonstrated impressive performance, surpassing other machine and deep learning-based research work. We have rigorously evaluated the effectiveness of our approach using two different metrics: Binary Accuracy and Macro F1 score.

There are some areas in this work that could be improved upon. In the future, we will address these issues by implementing a more efficient label encoding technique and reducing the amount of sparse data in a manner similar to what was done in this work. Additionally, we aim to optimize the method section of the HowSumm data, which contains more extensive sequences than the step data.



%
%
%



\end{document}